\documentclass{ifacconf}

\usepackage{makecell}
\usepackage{amsmath}
\usepackage{amssymb}

\usepackage{array}   
\usepackage{booktabs} 
\usepackage{fix-cm}
\usepackage{algorithm}
\usepackage{algpseudocode}
\usepackage{graphicx}      
\usepackage{natbib}        
\newtheorem{definition}{Definition}

\begin{document}
\begin{frontmatter}

\title{Neural Operator based Reinforcement Learning for Control of first-order PDEs with Spatially-Varying State Delay\thanksref{footnoteinfo}} 

\thanks[footnoteinfo]{The paper is supported by the National Natural Science Foundation of China (62173084,62403305), the Project of Science and Technology Commission of Shanghai Municipality, China (23ZR1401800, 22JC1401403).}

\author[First]{Jiaqi Hu} 
\author[First]{Jie Qi} 
\author[Second]{Jing Zhang}

\address[First]{College of Information Science and Technology, Donghua University Shanghai, China, (e-mail: hujiaqienjoy@gmail.com, jieqi@dhu.edu.cn).}
\address[Second]{College of Information Engineering, Shanghai Maritime University, Shanghai, China, (e-mail: zhang.jing@shmtu.edu.cn)}

\begin{abstract}
   Control of distributed parameter systems affected by delays is a challenging task, particularly when the delays depend on spatial variables. The idea of integrating analytical control theory with learning-based control within a unified control scheme is becoming increasingly promising and advantageous. In this paper, we address the problem of controlling an unstable first-order hyperbolic PDE with spatially-varying delays by combining PDE backstepping control strategies and deep reinforcement learning (RL). To  eliminate the assumption on the delay function required for the backstepping design, we propose a soft actor-critic (SAC) architecture incorporating a DeepONet to approximate the backstepping controller. The DeepONet extracts features from the backstepping controller and feeds them into the policy network. In simulations, our algorithm outperforms the baseline SAC without prior backstepping knowledge and the analytical controller.
   \end{abstract}


\end{frontmatter}
\footnotetext{This work has been submitted to IFAC for possible publication.}

\section{Introduction}
	We consider a first-order hyperbolic PDE with spatially-varying state delay
\begin{align}\label{eq:main-x0}
  \partial_t v(x,t)=& -\partial_x v (x,t)+ \int^1_{x} \! f(x,q)v(s,t)dq
 \nonumber\\ &+c(s)v(1,t-\tau(x)), 
  \\ \label{eq:bnd-x}
  v(0,t)=  &U(t), \\
  v(x,0)=&v_0(x),\\ 
  v(x,h)=&0, \qquad h\in [-\bar \tau,0),      \label{eq:x-before}
  \end{align}
  for $(x,t)\in (0,1) \times \mathbb{R}^{+}$. $\tau(x)$ is the delay dependent on the spatial argument $x$, with $\bar{\tau}=\sup_{x\in[0,1]}\tau(x)$.
  The backstepping controller for this system has been designed in \cite{zhang2021compensation, ZHANG2024105964} and its corresponding DeepONet-based controller is developed in \cite{qi2024nofeedback}. However, both controllers are limited by the assumption of slow variations in the delay i.e.,  $|\tau'(x)|<1$.  To remove the assumption, we propose a neural operator based Soft actor-critic (NO-SAC) architecture using a DeepONet approximating the backstepping controller and integrating it into the policy network. Therefore, this architecture utilizes the bacskstepping controller as prior knowledge and also take advantage of the adaptability and flexibility of the RL.


Classical control methods, such as the backstepping method \cite{krstic2008backstepping} and passivity-based control \cite{nuno2011passivity}, while theoretically precise and stability-guaranteeing, often encounter challenges like model mismatch or oversimplifications. These methods typically require stringent assumptions on the system coefficients.

On the other hand, data-driven methods, such as  RL \cite{schulman2017proximal, haarnoja2018soft}, learn control strategies directly through interaction with the environment, thus overcoming the dependency on accurate models (\cite{bhan2024pde,yu2022deep,mo2024proximal}).
While vanilla RL alone often suffers from slow convergence and stability issues. 
To address these shortcomings, recent studies begin to focus on integrating RL with prior theoretical knowledge.
Several studies have utilized Lyapunov stability theory to guide the design of RL algorithms, ensuring desirable stability of policies \cite{berkenkamp2017safe,chow2018lyapunov}. 
Other approaches focus on integrating theoretical insights into the RL to improve learning efficiency and stability. \cite{bougie2020towards} integrated prior knowledge and state similarity in the training. \cite{quartz2024stochastic} incorporated the linear quadratic regulator gain matrix computed around the steady-state operating point into the value function to further guide the policy. Besides, leveraging expert knowledge to reduce the dimensionality of the state space and constrain the action space also enhance learning efficiency and safety \cite{parisi2017goal, song2023reaching}.

Unlike traditional neural networks that operate on finite-dimensional spaces, DeepONet, as a neural operator (NO), excels in approximating high-dimensional mappings, making it ideal for PDE problems that couple spatial and temporal dimensions (\cite{lu2021learning}). The DeepONet leverages its unique branch and trunk network structure to learn operators directly and capture complex system behaviors with fewer data and computational resources. Meanwhile, the DeepONet is an effective approach for control problem, achieving a three-order-of-magnitude speedup in computation compared to traditional numerical methods. Examples are study from \cite{bhan2023neural,qi2024neural}, which shows an offline learning PDE backstepping control framework by the DeepONet. Their framework not only offers significant computational advantages but also achieves performance comparable to analyzed controllers, ensuring stability and control accuracy.

In this paper, we introduce a novel learning-based control approach by incorporating Neural Operators (NO) as a feature extractor within the Soft Actor-Critic (SAC) reinforcement learning (RL) framework, which we refer to as NO-SAC. First, we utilize the DeepONet to learn the backstepping controller for a first-order hyperbolic PDE with the spatially-varying delay. 
Second, the trained DeepONet is replicated and serves as a feature extraction network that is embedded into both the policy network and the value network. By leveraging prior backstepping knowledge through the DeepONet, the proposed NO-SAC framework demonstrates improved performance compared to the baseline SAC approach. The reward curve of NO-SAC over time consistently outperforms that of the baseline SAC. Additionally, the NO-SAC framework effectively eliminates the steady-state error in the closed-loop system, in contrast to the SAC. Additionally, we compare the control performance of the RL-based controller with that of the backstepping control method under the same delay function, which satisfies the delay assumption. The two RL-based control methods outperform the backstepping controller in terms of transient performance.

The primary contributions of this paper are as follows:
\begin{itemize}
  \item The proposed NO-SAC approach eliminates the assumption on the delay function that is necessary for backstepping control design.
  \item The DeepONet learns the backstepping prior knowledge and provides a warm start to both the actor and critic networks, leading to improved performance, particularly in eliminating steady-state error, when compared to the baseline SAC approach.
\end{itemize}

\section{DeepONet Learning Backstepping} \label{sec:2}
We introduce a two-dimensional transport PDE with spatially-varying speed to express the state delay in \eqref{eq:main-x0},
  \begin{align}
  	\partial_t v (x,t)&=-\partial_x v (x,t)+c(x)u(x,0,t) \nonumber\\ &\quad+\int^1_{x}\!  f(x,q)v(q,t)dq ,\label{eq:main-x1} \\
  	v(0,t)&=  U(t),\label{eq:bnd-x1}\\
  	\tau(x)\partial_t u (x,r,t)& = \partial_r u (x,r,t),~(x,r)\in (0,1)^2,
   \label{eq:main-u1}  \\
  	u(x,1,t)&=v(1,t), \label{eq:Combine-bnd-u1}\\
      v(x,0)&=  v_0(x),\label{eq:initial-x1}\\
  	u(x,r,0)&=u_0(x,r).   \label{eq:initial-u}
  \end{align} 
A strict assumption on the delay $\tau \in \mathcal{D} $ is necessary  for the bacsktepping control design, where
 \begin{align}
    \mathcal{D} =& \left\{ \tau \in C^2[0,1]: \tau(x) > 0~ ~\text{for} ~x\in [0,1] \right. \nonumber \\&\left.~~~~~\text{and ~if}~ \tau(x)<x, ~ |\tau'(x)| < 1  \right\}.\label{set:D}
  \end{align}
  

\begin{figure}[htb]
  \centering
  \includegraphics[width=8.4cm]{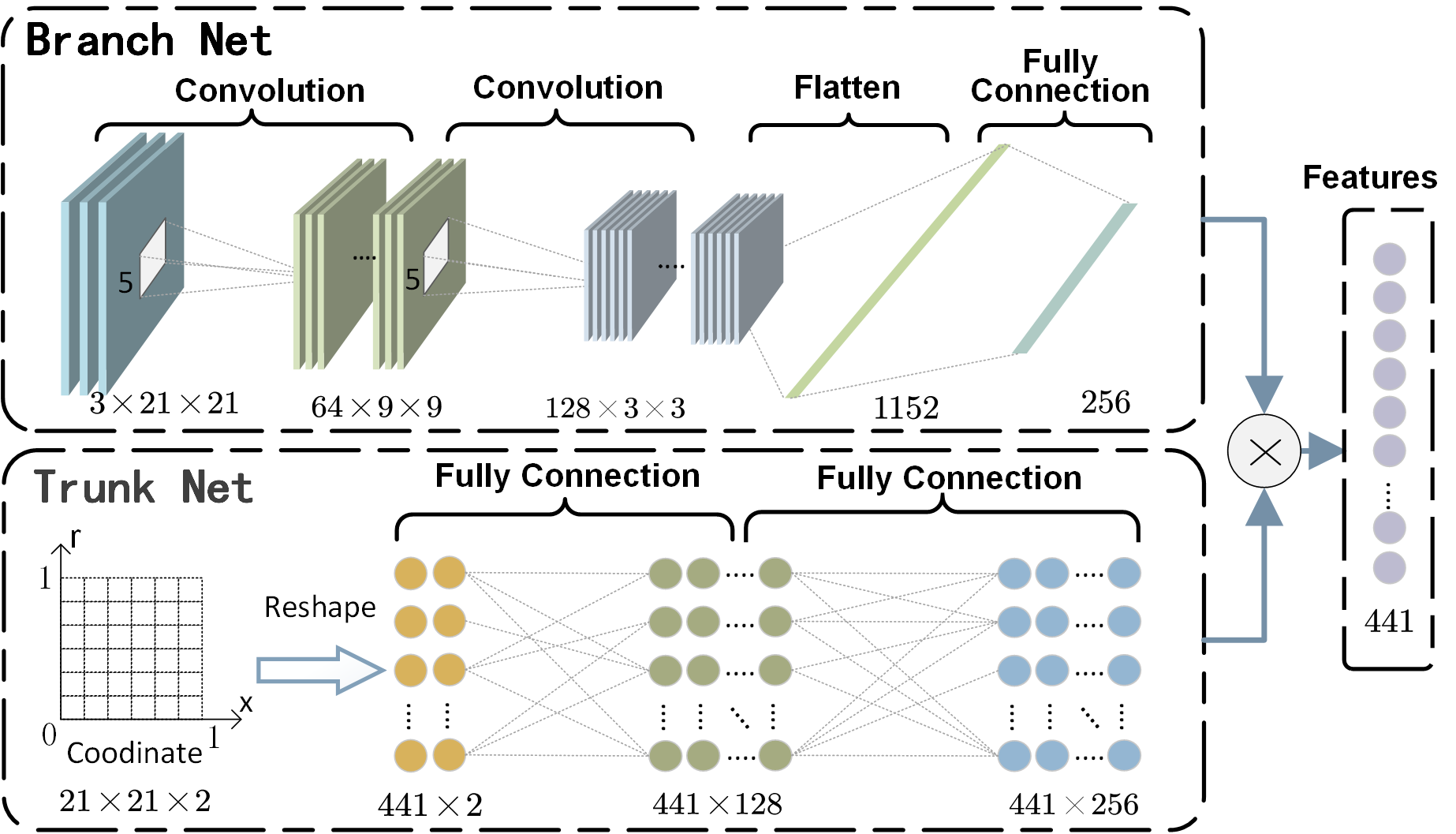}
  \caption{The structure of DeepONet.}
  \label{fig:deeponet}
\end{figure}

Define the backstepping controller designed in (\cite{zhang2021compensation, ZHANG2024105964}) as an operator
\begin{definition}\label{def:ctr_ope}
 The controller operator $\mathcal{U}: \mathcal{D}\times C^1[0,1]  
 \times C^1([0,1]^2)\mapsto \mathbb{R}$ with  	 
 \begin{align}\label{ope-U}
 U=\mathcal{U}( \tau, v,u),
 \end{align} 
 where  $\tau(x)$ is the delay function dependent on $x$,  $v(x, \cdot)$ and $u(x, r,  \cdot)$ are the system state and the delayed state, respectively. $U$ is the control input. 
\end{definition}

We apply a DeepONet to learn the controller operator \eqref{ope-U}. 
Therefore, the inputs to DeepONet, as shown in Fig.~\ref{fig:deeponet}, consist of these three components, i.e., $\tau, x,u$.  Unlike traditional neural networks that operate on finite-dimensional spaces, DeepONet is composed of a branch and trunk network structure. This architecture allows it to approximate operators and captures complex relationships in infinite-dimensional function spaces.

To match the domain of $(x, r)$, we discretize the 2-D domain $[0,1]^2$ spatially with a step size of $0.05$, resulting in 
 tensor input of size $3 \times 21 \times 21$ for the branch network. The branch network consists of two convolutional layers (kernel size $5 \times 5$, stride $2$) and a $1152 \times 256$ fully connected layer. The trunk network comprises two fully connected layers encoding inputs sampled on a $21 \times 21$ grid. 
The outputs of two networks are combined through a Cartesian product operation, producing a feature representation of size $441$.

Once the features of the backstepping controller is approximated by the DeepONet, we can embed the DeepONet into the SAC framework and use it to warmly start the RL-based controller. 

  \section{Neural Operator based Reinforcement Learning}
\begin{figure}[htb]
      \centering
      \includegraphics[width=8.4cm]{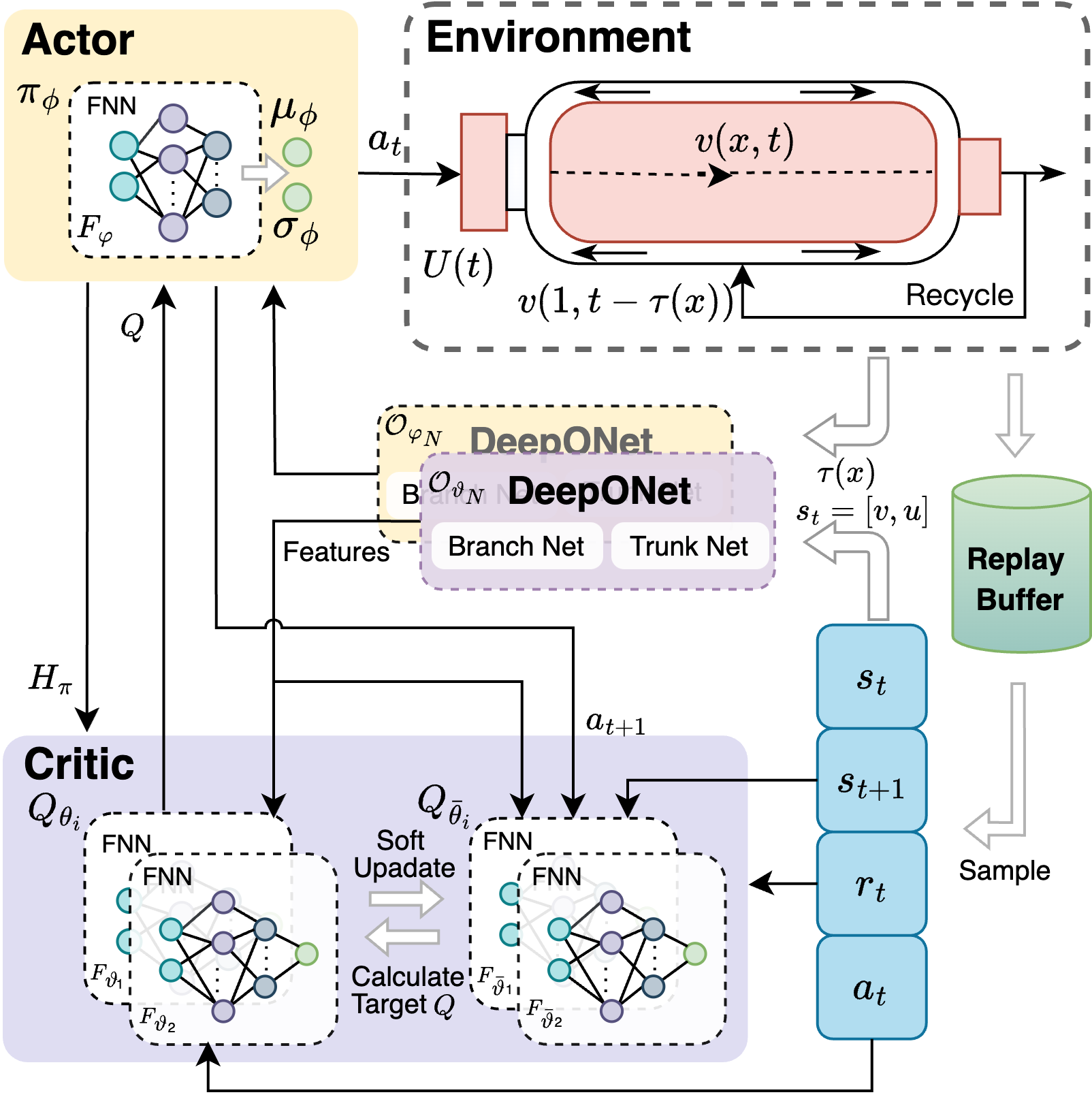}
      \caption{SAC architecture incorporating a DeepONet for 
    approximating the backstepping controller.}
      \label{fig:sketch}
    \end{figure}

    This section presents a RL framework that integrates the backstepping DeepONet with the SAC algorithm to control the delayed system.  In this framework, pre-training DeepONet in Fig.~\ref{fig:deeponet} serves as feature extractor for both the actor and critic networks, as illustrated in Fig.~\ref{fig:sketch}. The DeepONet processes the high-dimensional vectors composed of the states from the replay buffer and the system’s delay. Subsequently, the actor-critic networks are trained using the SAC algorithm.

    \subsection{Markov Decision Process for the delayed system}
    Due to the existence of the delay $\tau(x)$, the dynamics of the  system described by equations \eqref{eq:main-x1}-\eqref{eq:initial-u}  are non-Markovian \cite{bouteiller2020reinforcement}, as the next state $v(x, t + \Delta t)$ depends on both the current state $v(x, t)$ and its past state $v(1, t - \tau(x))$. We reformulate this system as a Markov Decision Process (MDP) in an augmented state space consisting of  $v(x, t)$ and $u(x,r,t)$, as defined in \eqref{eq:main-u1}-\eqref{eq:initial-u}.

  \textbf{State space} $\mathcal{S}$: State space is defined as  $\mathcal{S}=\{L^2[0,1]\times L^2[0,1]^2\}$. We denote $s_t=\{v(x,t), u(x,r,t)\}\in \mathcal{S}$ as the augmented state at time $t$ .

  \textbf{Action space} $\mathcal{A}\subseteq \mathbb{R}$: Defined as  $\mathcal{A} =[ -\overline{U},\overline{U}]$, where $\overline{U}=\text{sup}_{t\in \mathbb{R}^+}|U(t)|$. One can select an action $U(t)=a_t \in \mathcal{A}$ at each time step $t$ as the control input. 

\textbf{Policy function} $\pi(a_t|s_t)$: The policy is a probability density function $\pi(a_t|s_t)=\mathbb{P}(a_t | s_t)$ that maps the current state $s_t \in \mathcal{S}$ to a probability distribution over actions $a_t \in \mathcal{A}$. In our framework, controller is paramterized by a neural network as policy, denoted by $\pi_{\phi}$, which  is assumed to belong  a Gaussian distribution family
\begin{align}
    \pi_{\phi}\in \{\mathcal{N}(\mu_\phi, \sigma_\phi)\}
\end{align}
 where $\mu_\phi$ and $\sigma_\phi$ are the mean and standard deviation.

  \textbf{State transition $p(s_{t+1}|s_t,a_t)$}: By augmenting the state with delay information $u(s,r,t)$, the transition dynamics become Markovian. The probability $\mathbb{P}(s_{t+1}|s_t, a_t)$ depends on the augmented state $s_t$ and the selected action $a_t$. The system evolves according to the dynamics described in \eqref{eq:main-x1}--\eqref {eq:initial-u}, where the state $s_t$ transitions to $s_{t+1}$ under the influence of the action $a_t \in \mathcal{A}$.

\textbf{Reward function} $r_t$: It is designed to guide the learning of an optimal control policy and comprises two components,
\begin{align}\label{eq:reward_total}
    r_t = r_{\text{mid}} + r_{\text{end}},
\end{align}
where 
\begin{align}
    r_{\text{mid}} =& - \ln\left(1+\|s_{t-1} - s_t \|_{L_2} \right) \\\notag&- \Gamma \cdot\ln\left(1+\|s_t\|_{L_2} \right),\\
    \label{eq:reward_end}
      r_{\text{end}} =& 
    \begin{cases} 
    0, & \|s_T\|_{L_2} > \zeta,\\ 
    \frac{ \sigma}{1+\sum_{i = 0}^T \|s_i\|_{L_2}},   & \|s_T \| _{L_2} \leq \zeta.
    \end{cases}
  \end{align}
where $\Gamma>0$ denotes weighting factor, $T$ denotes final step of each episode, $\zeta$ represents the threshold for the $L_2$ norm of the final state $s_T$, and $\sigma$ is scaling factor that adjusts reward magnitude. The second component provides additional reward if state approaches the equilibrium point $\mathbf{0}$.

\textbf{Return} $R_t$:  It denoted as $R_t = \sum_{k=t}^{\infty} \gamma^{k-t} r_k$, represents the discounted cumulative reward starting from time $t$ where $\gamma \in (0,1)$ .

\subsection{NO-SAC Control Design}
We employ the actot-critic SAC framework, which consists of an actor network $\pi_{\phi}(a_t|s_t)$, two critic networks $Q_{\theta_i}(s_t, a_t)$ and their target network $Q_{\bar{\theta}_i}(s_t, a_t)$, to avoid bootstrapping.  

The backstepping DeepONet is duplicated into five copies, in addition to the original one, which are incorporated into the actor network and critic networks. In this case, the actor network $\pi_\phi$ consists of the DeepONet $\mathcal{O}_{\varphi_N}$ and a fully connected neural network (FNN) $F_\varphi$, i.e., $\pi_\phi=\mathcal{O}_{\varphi_N} \cup F_\varphi$. Similarly, critic networks $Q_{\theta_i}$ consist of the duplicated DeepONets $\mathcal{O}_{\vartheta_{N}}$ and  FNNs  $F_{\vartheta_i}$, i.e., $Q_{\theta_i}=\mathcal{O}_{\vartheta_{N}} \cup F_{\vartheta_i}$ for $i=1,2$.  

The actor and critic networks are updated via backpropagation, and their respective feature extraction layers $\mathcal{O}_{\varphi}$ for the actor and $\mathcal{O}_{\vartheta_i}$ for the critic networks, are optimized independently. 



\textbf{Learning critic networks}: Define action-value function as $Q_\pi (s_t,a_t)=Q: \mathcal{S}\times\mathcal{A}\mapsto \mathbb{R}$ representing the expected return following policy $\pi_{\phi}$ after taking action $a_t$ in state $s_t$.
To reduce bootstrapping bias, the SAC employs two target networks $Q_{\bar\theta_1}$ and $Q_{\bar\theta_2}$ to estimate the target action-value
\begin{align}\label{eq:tar_q}
y = r_t + \gamma \left( \min_{i=1,2} Q_{\bar{\theta}_i}(s_{t+1}, \breve a_{t+1}) - \alpha \log \pi_\phi(\breve a_{t+1}|s_{t+1}) \right),
\end{align}
where $\alpha$ is the temperature parameter that adjust the trade-off between reward maximization and entropy, and the actions $\breve a_{t+1}\sim \pi(\cdot|s_{t+1})$ is generated from the policy network.

The action-value networks, $Q_{{\theta}_1}(s_t, a_t)$ and $Q_{\theta_2}(s_t, a_t)$, are trained by minimizing the TD soft residual
\begin{equation}
J(Q_{\theta_i}) = \mathbb{E}_{(s_t, a_t) \sim \mathcal{W}} \left[ \left( Q_{\theta_i}(s_t, a_t) - y \right)^2 \right],
\end{equation}
where $i \in {1, 2}$ and $\mathcal{W}$ is the distribution of the states and the actions.

The weights of the two target action-value networks $Q_{\bar\theta_1}$ and $Q_{\bar\theta_2}$  are updated by
\begin{equation}\label{eq:up_tar_q}
\bar{\theta}_i \leftarrow \eta \theta_i + (1 - \eta) \bar{\theta}_i,
\end{equation}
where $i \in {1, 2}$ and $\eta\in (0, 1]$ is the weighted coefficient.

\textbf{Learning the Policy}: 
\begin{definition}\label{def:entropy}
    (Entropy)  The entropy $H(\pi(\cdot|s_t))$ of the policy  measures the randomness of the action distribution given a state $s_t$, defined as:
\begin{align}
H(\pi(\cdot|s_t)) = -\mathbb{E}_{a_t \sim \pi_\phi} \left[\log \pi_\phi(a_t|s_t)\right],
\end{align}
\end{definition}
The policy network $\pi_\phi(a_t|s_t)$ is trained by minimizing the following objective:
\begin{align}
J(\pi_\phi) = \mathbb{E}_{s_t \sim \mathcal{G}} \mathbb{E}_{a_t \sim \pi_\phi} \left[- \alpha H(\pi_{\phi}(\cdot|s_t)) - \min_{i=1,2} Q_{\theta_i}(s_t, a_t) \right],
\end{align}
where $\mathcal{G}$ denotes  the state distribution.

\begin{algorithm}
\caption{Soft Actor-Critic (SAC) Procedure}
\begin{algorithmic}[1]
\State Initialization: $\phi$, $\theta_i$, $\bar{\theta}_i$ for $i=1,2$, $\mathcal{W}_s$ and $s_0=\{v_0, 0\}$
\For{$l \gets 0$ \textbf{to} total steps}
    \State Observe state $s_t=\{v_t, u_t\}$ 
    \State Select action $a_t \sim \pi_\phi(a_t|s_t)$

    \State Execute action $a_t$ in environment, observe $s_{t+1}$, $r_t$
    \State $\mathcal{W}\gets \mathcal{W}\cup\{(s_t, a_t, r_t, s_{t+1})\}$ 
    \If{$\|s_t\|_{L_2}$ exceeds limit  \textbf{or} $t$ reaches terminal time}
        \State Reset initial state $s_0$ 
    \EndIf
    \For{each gradient step}
    \State Sample a minibatch of transitions from $\mathcal{W}$
    \State Compute target value $y$ by equation \eqref{eq:tar_q}
    \State Update action-value network parameters $\theta_i=\{\vartheta_N, \vartheta_i\}$: $Q_i\gets Q_i-\lambda \hat{\nabla}_{Q_i} J(Q_i)$
    \State Update actor network parameters $\phi=\{\varphi_N, \varphi\}$: $\pi \gets \pi-\lambda \hat{\nabla}_{\pi} J(\pi)$
    \State Update target critic Networks by equation \eqref{eq:up_tar_q}
\EndFor
\EndFor

\end{algorithmic}
\end{algorithm}


After training is complete, the spatially discretized variables $v(x,\cdot)$ and $u(x,r,\cdot)$ and $\tau(x)$, along with their corresponding discrete coordinates in $[0,1]^2$, are fed into the NO-SAC, which generates the mean $\mu_\phi$. We use it as the control input.


\section{Simulation Results}
In this section, we design a delay compensate controller based on the SAC algorithm and DeepONet, bypassing the constraints of the backstepping method that required specific assumptions about the delay, i.e., $\tau \in \mathcal{D}$. The DeepONet shown in Fig.~\ref{fig:deeponet} serves as the feature extraction layer of the Actor-Critic network, enabling the extraction of features from high-dimensional data based on the backstepping controller. The extracted features, with a shape of 441, are further processed by the Actor-Critic network to output action values $Q(s_t,a_t)$ and control $U(t)$.


\subsection{Configuration}
The training process was conducted on a workstation equipped with an Intel Core i9-13900KF CPU and an NVIDIA GeForce RTX 4090 GPU. We construct a Gym environment based on equations \eqref{eq:main-x0}-\eqref{eq:x-before}, following \cite{bhan2024pde}. The parameters of reward \eqref{eq:reward_total} are set as $\Gamma=0.008$, $\sigma=300$, and $\zeta=10$. Built upon the Stable-Baselines3 library (SB3), the framework used a replay buffer size of $10^5$, with discount reward factor $\gamma=0.99$. Key hyperparameters are selected as follows: learning rate $\lambda=0.00009$, policy update frequency of 2, and soft update coefficient $\eta=0.003$. Default SB3 parameters are omitted for brevity.


\begin{figure}[htb]
  \centering
  \includegraphics[width=7.4cm]{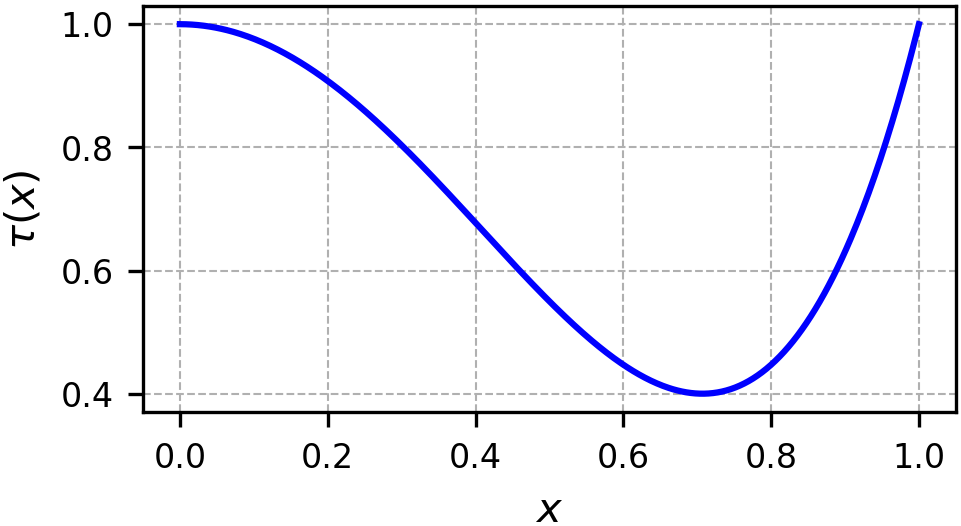}
  \caption{Delay function $\tau(x)=0.7+0.3\cos(4\arccos(x))$.}
  \label{fig:delay_curve}
\end{figure}


Each episode lasts 5 seconds and consists of 2500 steps under a temporal step size setting of $\Delta t = 0.002$. The training process comprises 100 episodes in total taking approximately 40 minutes. 
For plant coefficient, the attenuation factor $c(x)=20(1-x)$ and heat transfer coefficient $f(x,q)=5\cos(2\pi q)+5\sin(2\pi x)$ are used. We use the delay function $\tau(x)=0.7+0.3\cos(4\arccos(x))$, which violates the assumption $\tau\in \mathcal{D}$, where $\mathcal{D}$ is defined in \eqref{set:D}, as shown in Fig.~\ref{fig:delay_curve}. During training, the initial state $v_0(x)$ is a constant randomly choosing from a uniform distribution $U[1,8]$ and let $u_0(x,r)=0$. For testing, the initial state is set to $v_0(x) = 6$. The state space for interaction is defined as  $[-\infty, \infty]$, and the action space is constrained to $[-30, 30]$.

In real-world applications, controller frequencies are typically limited to 100 Hz due to sensor and actuator response times. Excessively high frequencies can hinder the RL process \cite{bhan2024pde}, causing instability or learning difficulties. In our framework, we update the control strategy every $100$ steps using a zero-order hold.



\begin{figure}[htb]
  \centering
  \includegraphics[width=7.5cm]{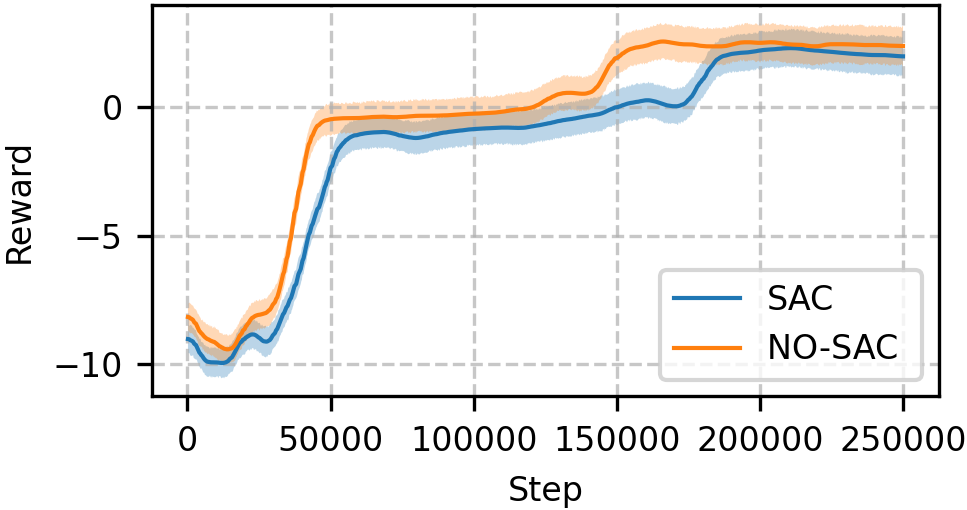}
  \caption{The reward  evolution.}
  \label{fig:reward_curve}
\end{figure}


\subsection{Simulation Results}
To evaluate the performance of our framework, we established a baseline SAC (\cite{haarnoja2018soft}) for comparison. Reward evolution curves for the baseline SAC and the proposed algorithm, NO-SAC, are shown in Fig.~\ref{fig:reward_curve}. The curves illustrate that NO-SAC converges faster and consistently outperforms the baseline in terms of reward. It is noteworthy that the reward curves does not converge to zero, due to the positive reward $r_{\text{end}}$ defined in equation \eqref{eq:reward_end}. This reward definition adds additional rewards when the states converge to the expected values.

\begin{figure*}
    \centering
        \includegraphics[width=5.4cm]{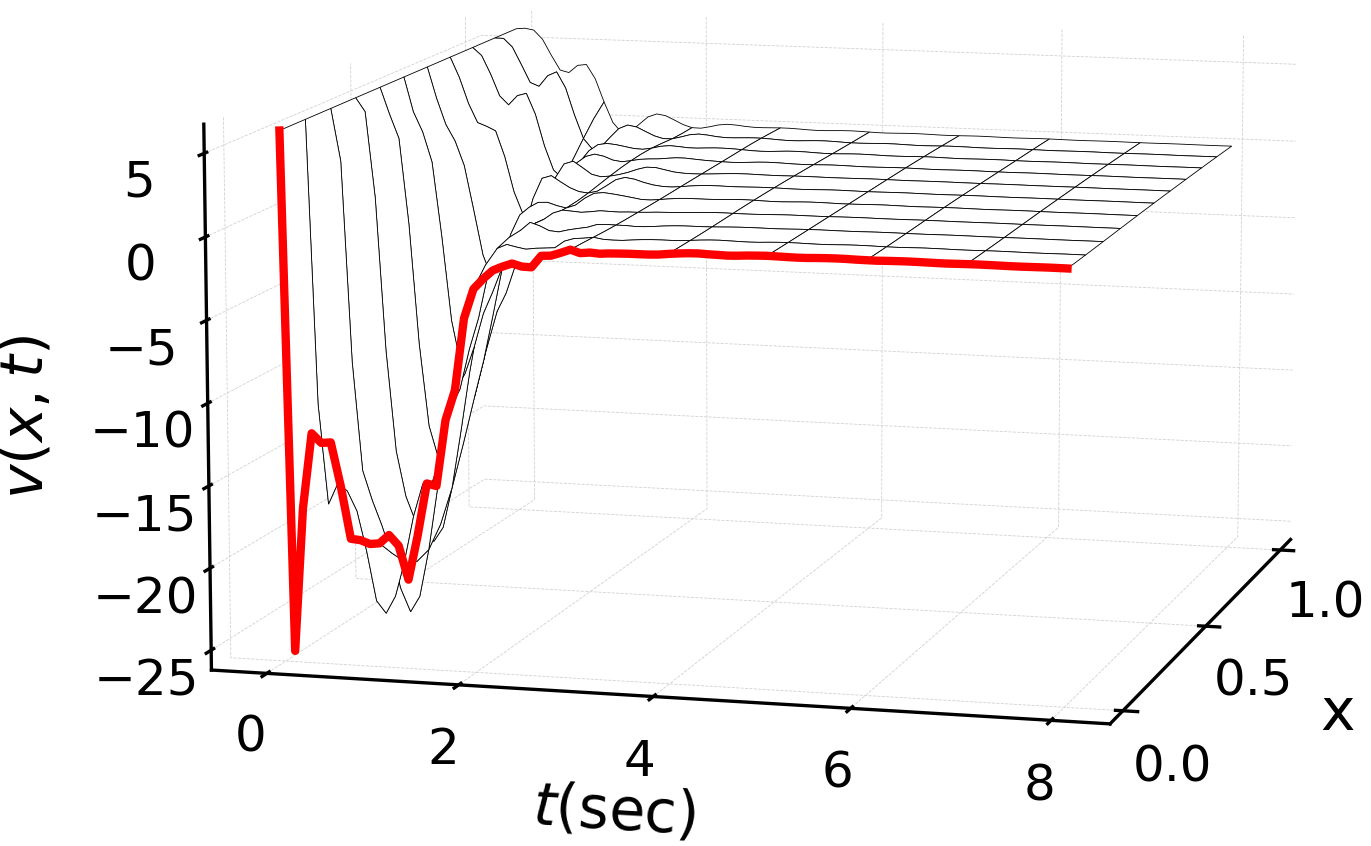}
        \includegraphics[width=5.4cm]{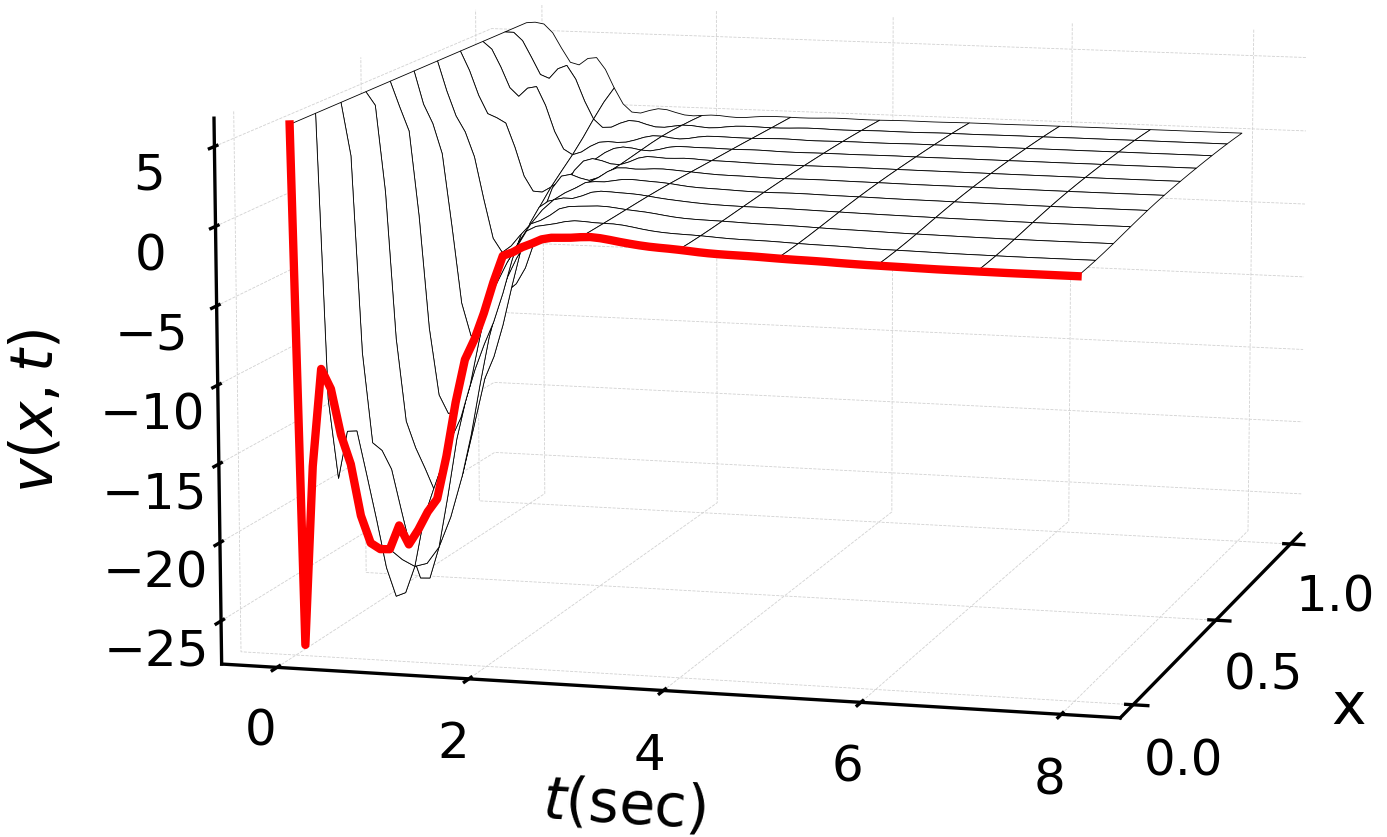}
    \caption{Close-loop state $v(x,t)$ with initial condition $v_0=6$ and delay $\tau(x) = 0.7+0.3\cos(\arccos(x))$ using the NO-SAC (left) and the SAC (right).}
    \label{fig:result_no_limit}
\end{figure*}

\begin{figure*}[htb]
  \centering
  \includegraphics[width=7.8cm]{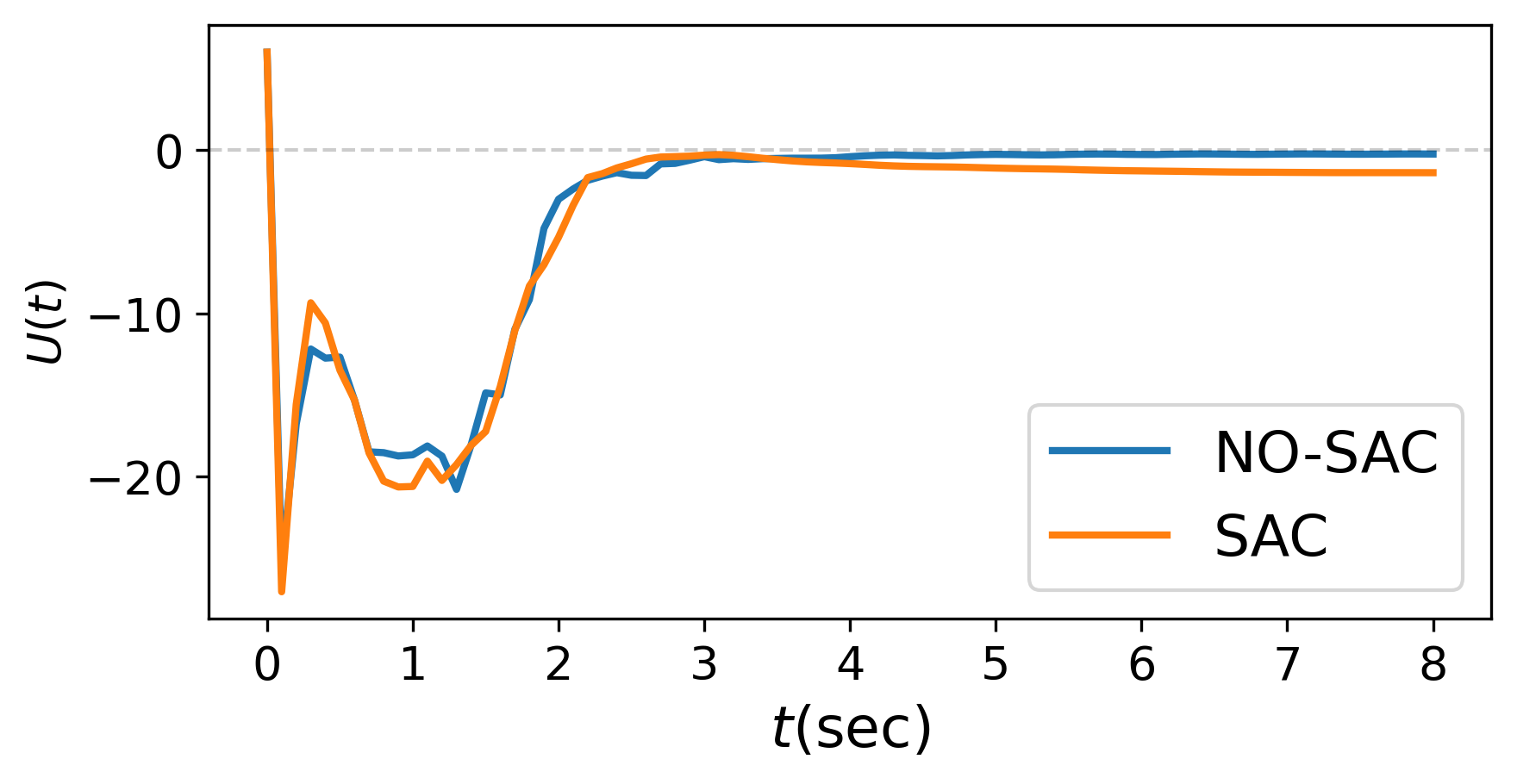}
\includegraphics[width=7.8cm]{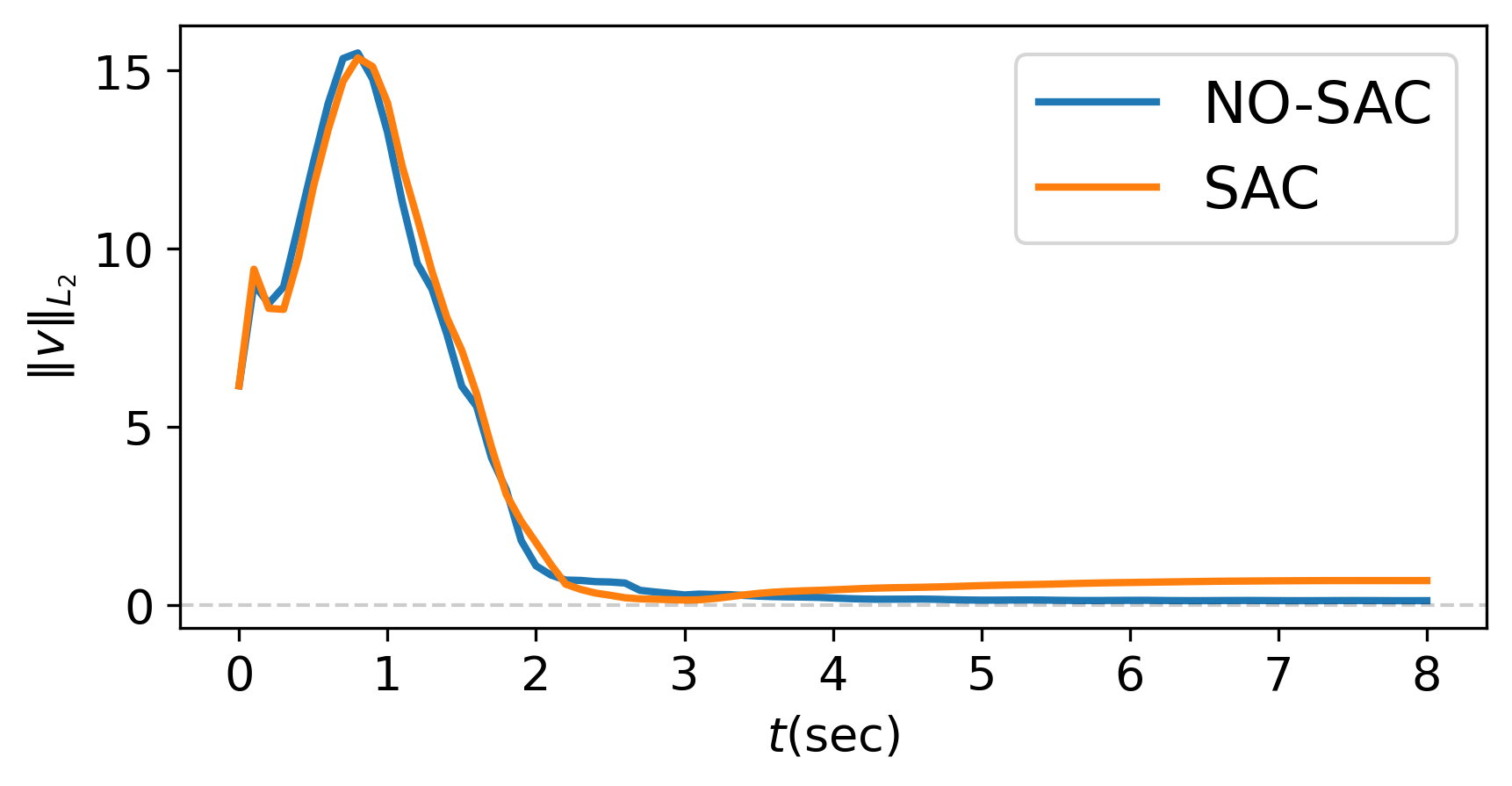}\\
  \includegraphics[width=7.8cm]{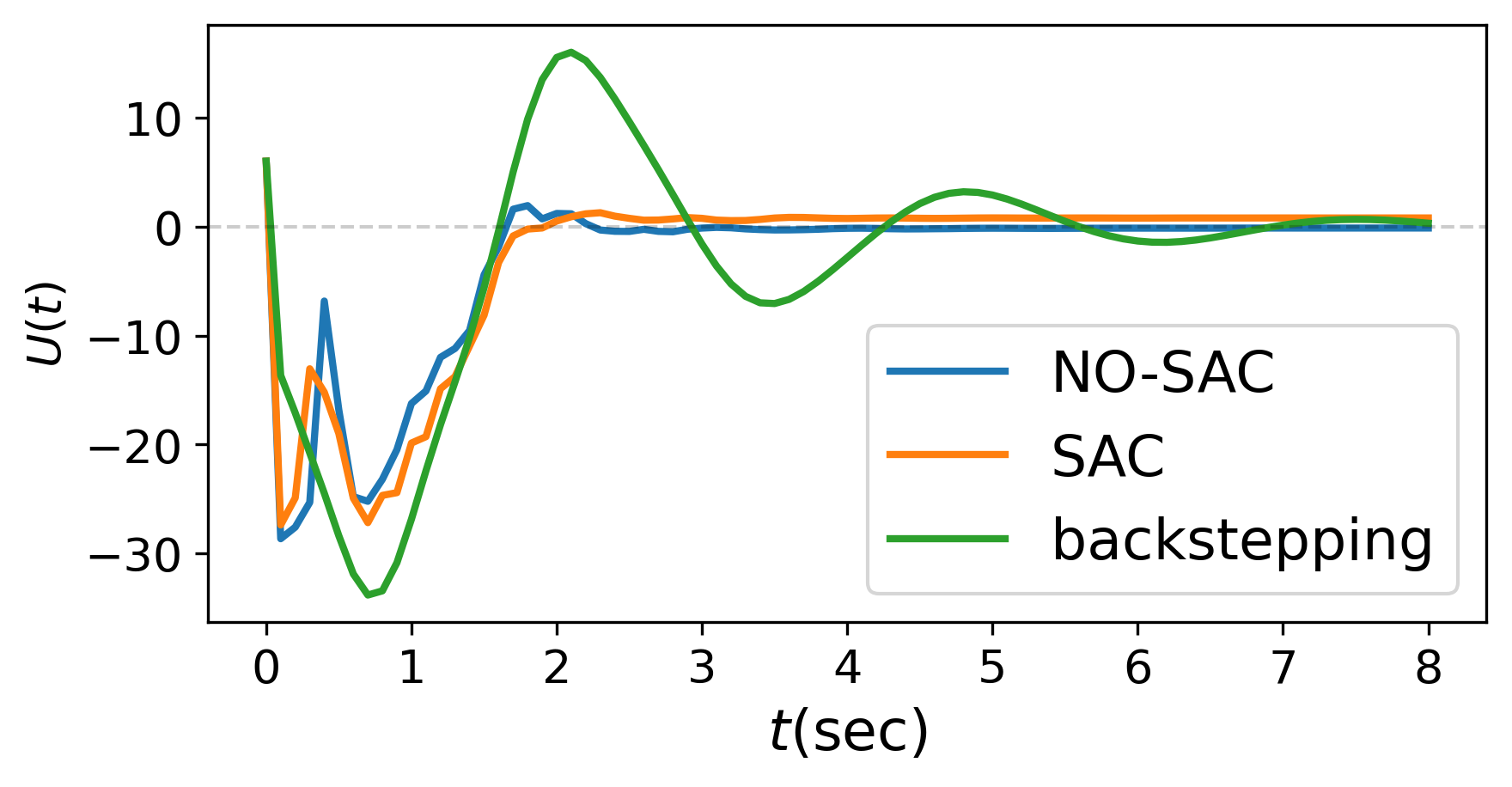}
    \includegraphics[width=7.8cm]{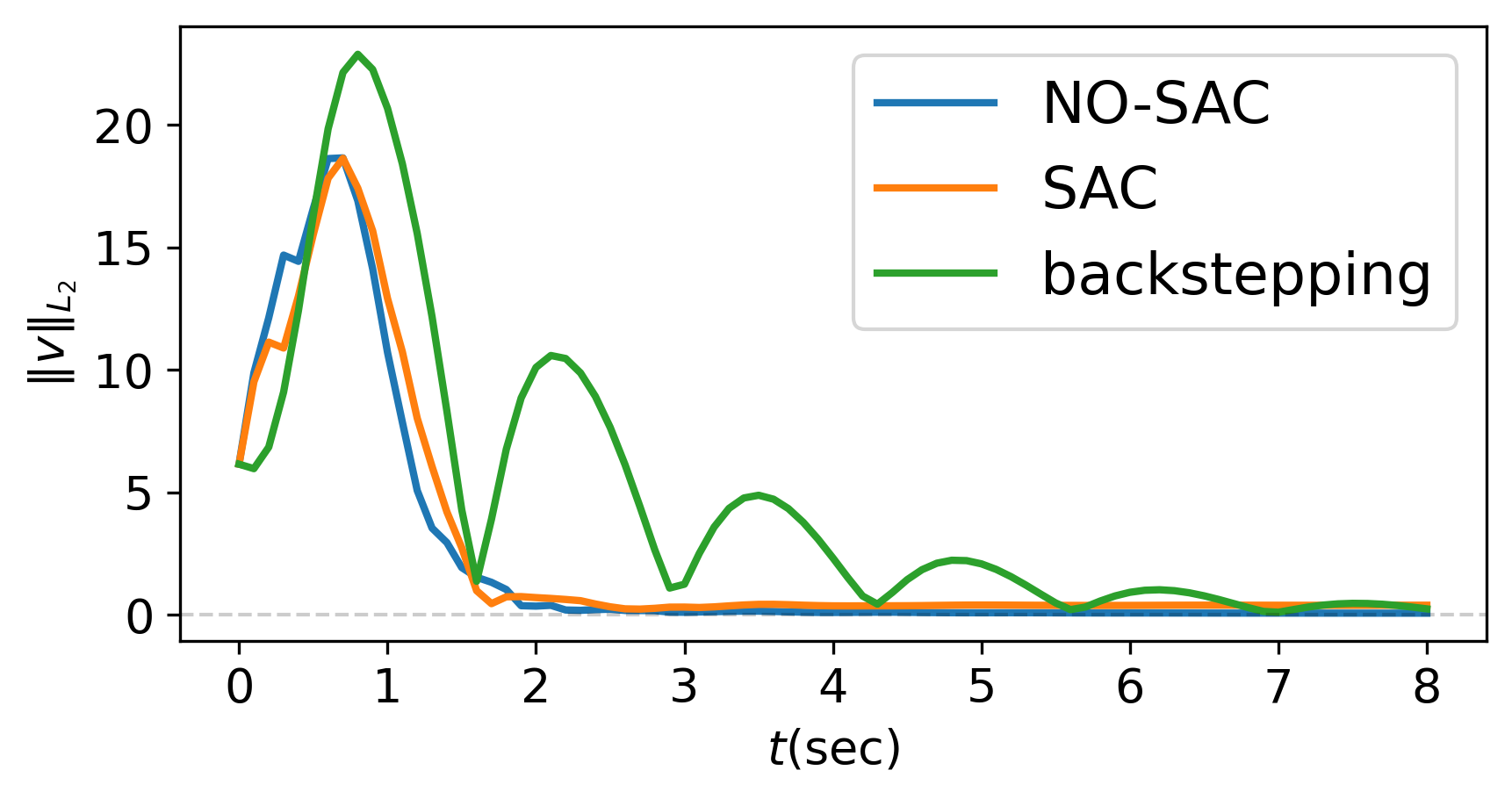}
  \caption{Control input $U(t)$ (Left), and $L_2$ norm of state $v(x,\cdot)$ (Right). The top and bottom panels correspond to $\tau(x) = 0.7+0.3\cos(\arccos(x))$ and $\tau(x) = e^{-0.7x}$, respectively.}
  \label{fig:compare}
\end{figure*}

Fig.\ref{fig:result_no_limit} illustrates the closed-loop state evolution using the NO-SAC controllers and the SAC whitout considering the assumption of $\tau \in \mathcal{D}$. To compare the RL-based controllers with the backstepping controller, we plot the closed-loop state evolution with three controllers for $\tau(x) = e^{-0.7x}$ satisfying the delay assumption, shown in Fig.\ref{fig:result_no_limit}. In addition, Fig.\ref{fig:compare} illustrates the control inputs of different controllers and the $L_2$ norm of the state for different $\tau(x)$. 


\begin{figure*}[htb]
  \centering
    \includegraphics[width=5.4cm]{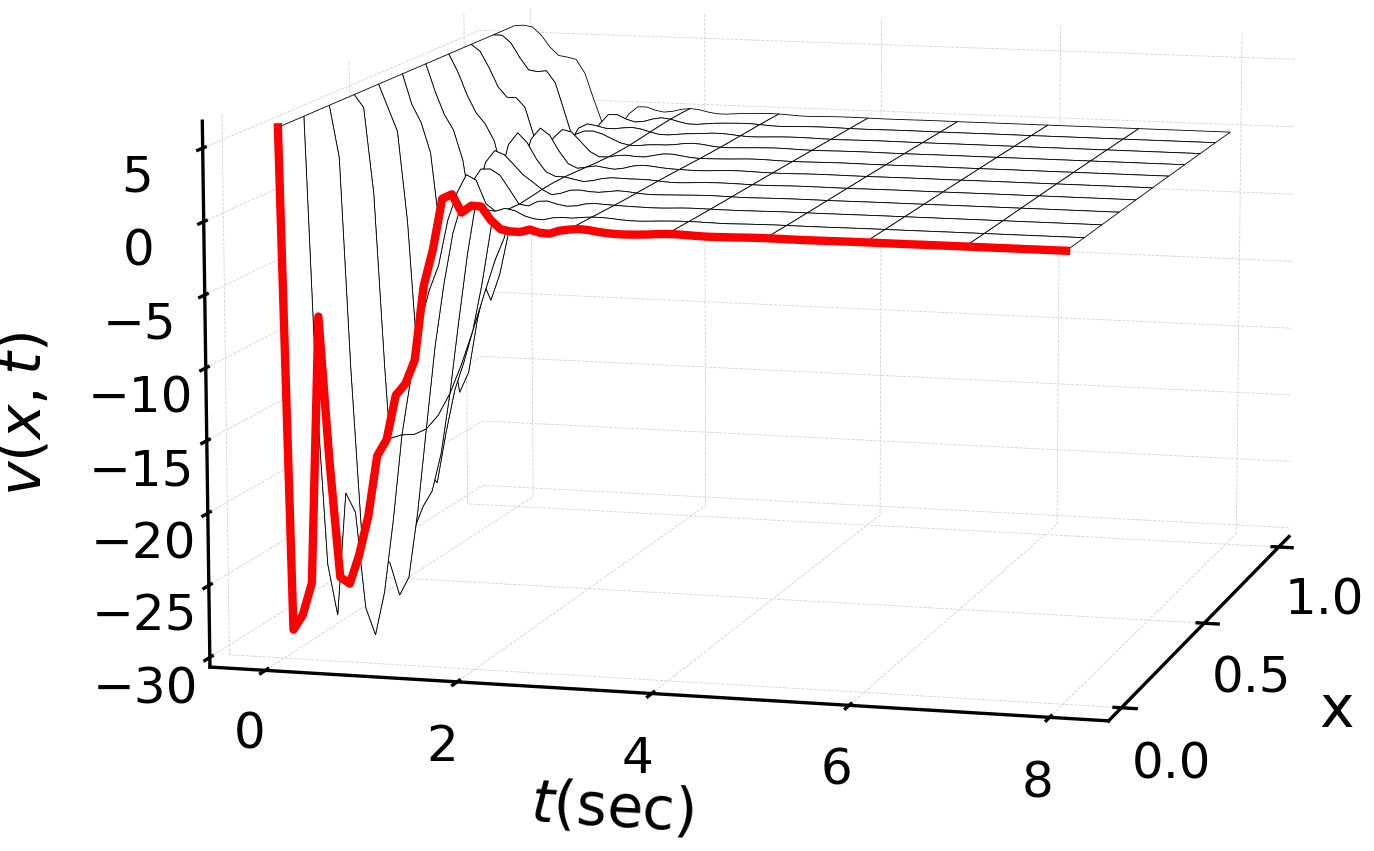}
    \includegraphics[width=5.4cm]{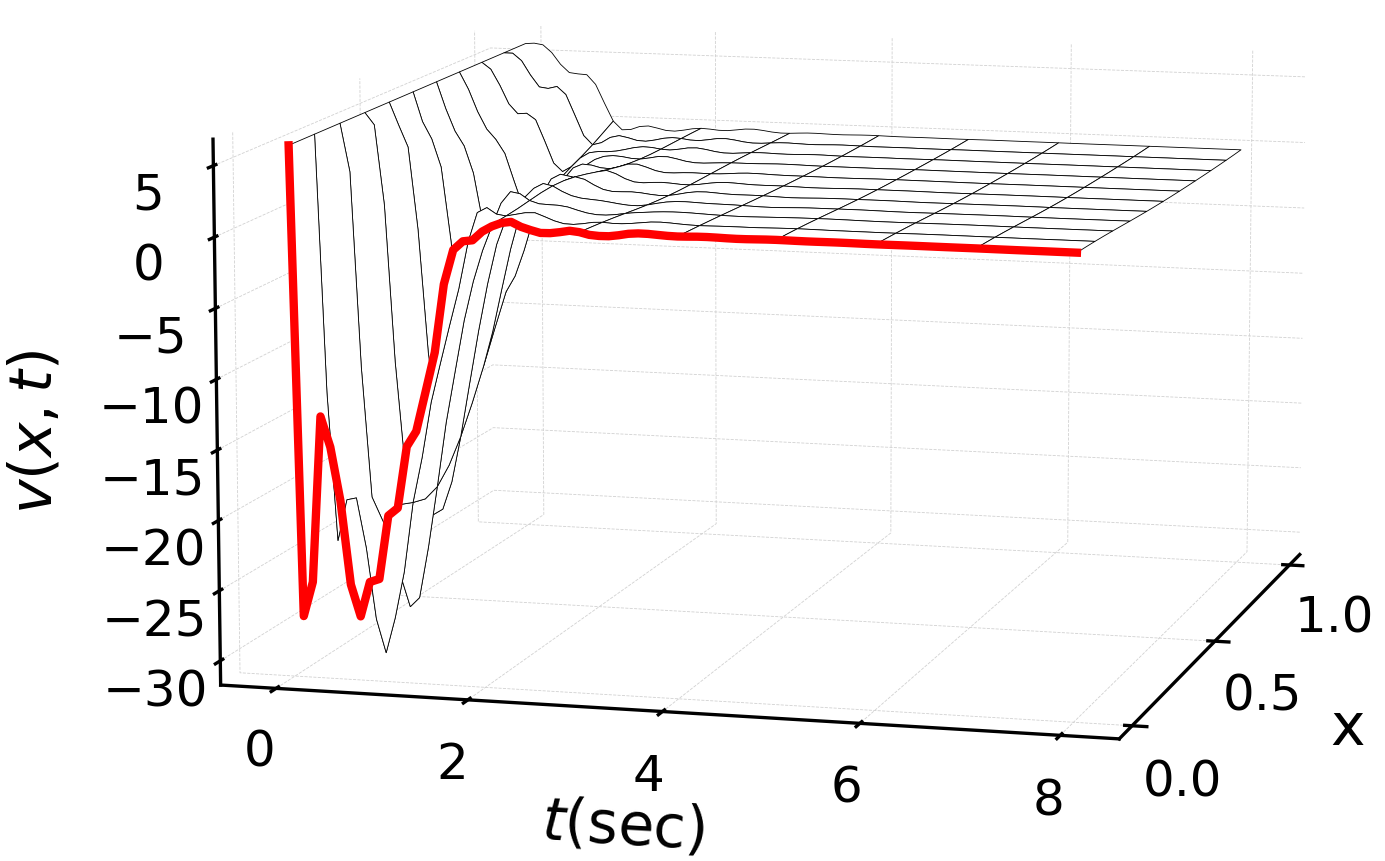}
    \includegraphics[width=5.4cm]{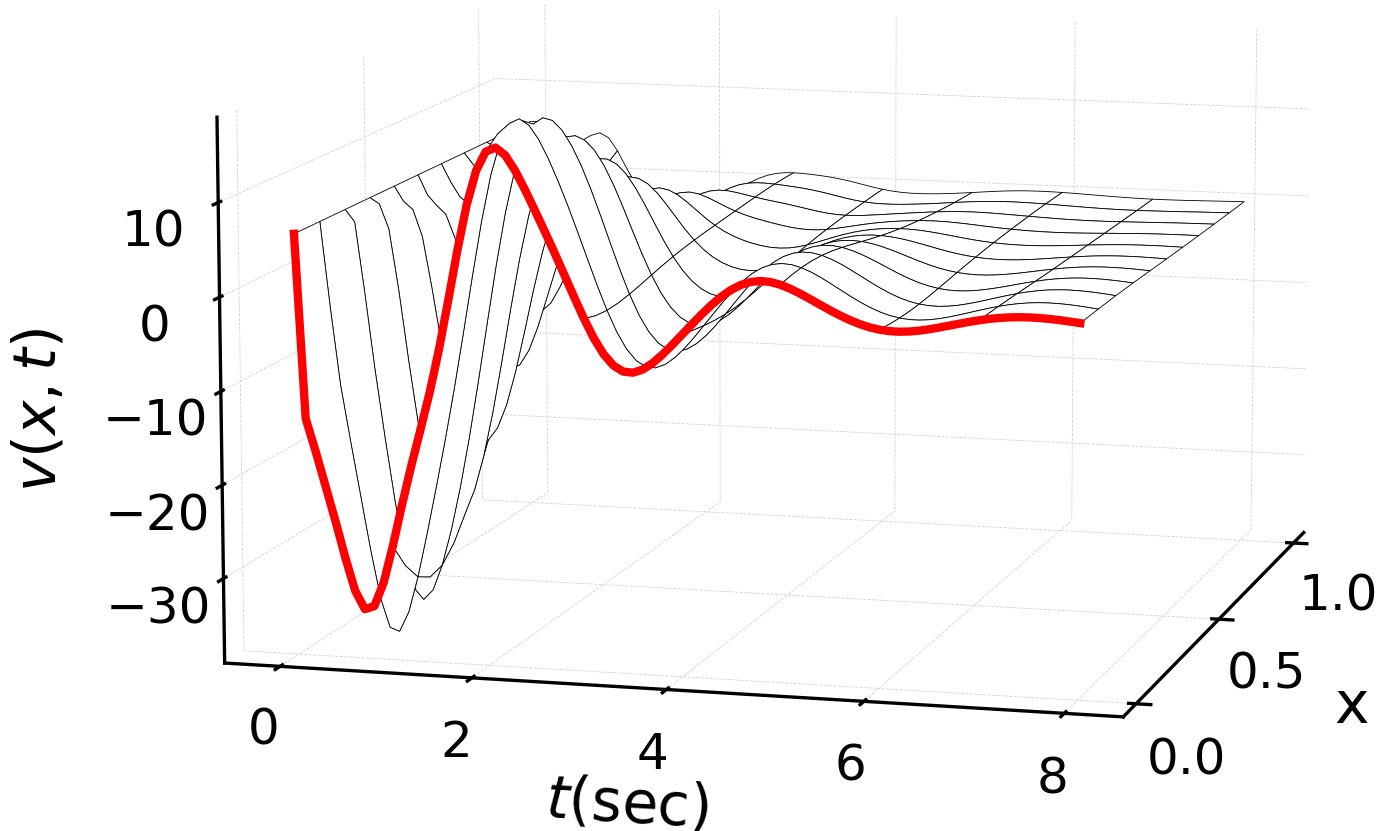}
  \caption{Close-loop state $v(x,t)$ with initial condition $v_0=6$ and delay function $\tau(x) = e^{-0.7x}$ using the NO-SAC controller (Left), the SAC controller (Middle) and the backsteping controller (Right).}
  \label{fig:result}
\end{figure*}

The simulation results demonstrate that incorporating a pre-trained DeepONet as the feature extraction layer within the SAC algorithm increases reward acquisition and improves training efficiency. Furthermore, the NO-SAC-based control strategy reduces steady-state error compared to the SAC-based control strategy. From Fig. \ref{fig:compare}, we observe that the RL-based controllers exhibit smaller overshoot and shorter settling times compared to the backstepping controller under the same delay function.

\section{Conclusion}
In this paper, we propose a NO-SAC framework, which eliminates the assumptions on the delay function in the backstepping design by integrating backstepping control strategies with the RL and stabilizes an unstable first-order hyperbolic PDE with spatially varying delays. Our framework integrates the backstepping DeepONet as feature extractor within the SAC algorithm. This hybrid approach achieves faster convergence and decreases the transient overshoot compared to the backstepping controller.
From the perspective of RL, NO-SAC eliminates steady-state errors in dynamics and demonstrates superior generalization performance compared to the SAC. These improvements stem from DeepONet, which maps infinite-dimensional function spaces for more efficient feature extraction and better generalization. Future work will advance the theory of operator learning in control, extending beyond DeepONet to develop more generalizable representations and analyze their impact on stability in RL-based controllers for distributed systems.


\bibliography{reference_paper.bib}         

\begin{thebibliography}{20}
\providecommand{\natexlab}[1]{#1}
\providecommand{\url}[1]{\texttt{#1}}
\providecommand{\urlprefix}{URL }
\expandafter\ifx\csname urlstyle\endcsname\relax
  \providecommand{\doi}[1]{doi:\discretionary{}{}{}#1}\else
  \providecommand{\doi}{doi:\discretionary{}{}{}\begingroup \urlstyle{rm}\Url}\fi

\bibitem[{Berkenkamp et~al.(2017)Berkenkamp, Turchetta, Schoellig, and Krause}]{berkenkamp2017safe}
Berkenkamp, F., Turchetta, M., Schoellig, A., and Krause, A. (2017).
\newblock Safe model-based reinforcement learning with stability guarantees.
\newblock \emph{Advances in neural information processing systems}, 30.

\bibitem[{Bhan et~al.(2024)Bhan, Bian, Krstic, and Shi}]{bhan2024pde}
Bhan, L., Bian, Y., Krstic, M., and Shi, Y. (2024).
\newblock {PDE} control gym: A benchmark for data-driven boundary control of partial differential equations.
\newblock \emph{arXiv preprint arXiv:2405.11401}.

\bibitem[{Bhan et~al.(2023)Bhan, Shi, and Krstic}]{bhan2023neural}
Bhan, L., Shi, Y., and Krstic, M. (2023).
\newblock Neural operators for bypassing gain and control computations in {PDE} backstepping.
\newblock \emph{IEEE Transactions on Automatic Control}.

\bibitem[{Bougie and Ichise(2020)}]{bougie2020towards}
Bougie, N. and Ichise, R. (2020).
\newblock Towards interpretable reinforcement learning with state abstraction driven by external knowledge.
\newblock \emph{IEICE TRANSACTIONS on Information and Systems}, 103(10), 2143--2153.

\bibitem[{Bouteiller et~al.(2020)Bouteiller, Ramstedt, Beltrame, Pal, and Binas}]{bouteiller2020reinforcement}
Bouteiller, Y., Ramstedt, S., Beltrame, G., Pal, C., and Binas, J. (2020).
\newblock Reinforcement learning with random delays.
\newblock In \emph{International conference on learning representations}.

\bibitem[{Chow et~al.(2018)Chow, Nachum, Duenez-Guzman, and Ghavamzadeh}]{chow2018lyapunov}
Chow, Y., Nachum, O., Duenez-Guzman, E., and Ghavamzadeh, M. (2018).
\newblock A lyapunov-based approach to safe reinforcement learning.
\newblock \emph{Advances in neural information processing systems}, 31.

\bibitem[{Haarnoja et~al.(2018)Haarnoja, Zhou, Abbeel, and Levine}]{haarnoja2018soft}
Haarnoja, T., Zhou, A., Abbeel, P., and Levine, S. (2018).
\newblock Soft actor-critic: {Off-policy} maximum entropy deep reinforcement learning with a stochastic actor.
\newblock In \emph{International conference on machine learning}, 1861--1870.

\bibitem[{Krstic and Smyshlyaev(2008)}]{krstic2008backstepping}
Krstic, M. and Smyshlyaev, A. (2008).
\newblock Backstepping boundary control for first-order hyperbolic {PDEs} and application to systems with actuator and sensor delays.
\newblock \emph{Systems \& Control Letters}, 57(9), 750--758.

\bibitem[{Lu et~al.(2021)Lu, Jin, Pang, Zhang, and Karniadakis}]{lu2021learning}
Lu, L., Jin, P., Pang, G., Zhang, Z., and Karniadakis, G.E. (2021).
\newblock Learning nonlinear operators via {DeepONet} based on the universal approximation theorem of operators.
\newblock \emph{Nature machine intelligence}, 3(3), 218--229.

\bibitem[{Mo et~al.(2024)Mo, Wu, Qi, Pan, Feng, Yan, and Wang}]{mo2024proximal}
Mo, S., Wu, N., Qi, J., Pan, A., Feng, Z., Yan, H., and Wang, Y. (2024).
\newblock Proximal policy optimization learning based control of congested freeway traffic.
\newblock \emph{Optimal Control Applications and Methods}, 45(2), 719--736.

\bibitem[{Nu{\~n}o et~al.(2011)Nu{\~n}o, Basa{\~n}ez, and Ortega}]{nuno2011passivity}
Nu{\~n}o, E., Basa{\~n}ez, L., and Ortega, R. (2011).
\newblock Passivity-based control for bilateral teleoperation: A tutorial.
\newblock \emph{Automatica}, 47(3), 485--495.

\bibitem[{Parisi et~al.(2017)Parisi, Ramstedt, and Peters}]{parisi2017goal}
Parisi, S., Ramstedt, S., and Peters, J. (2017).
\newblock Goal-driven dimensionality reduction for reinforcement learning.
\newblock In \emph{2017 IEEE/RSJ International Conference on Intelligent Robots and Systems (IROS)}, 4634--4639. IEEE.

\bibitem[{Qi et~al.(2024{\natexlab{a}})Qi, Hu, Zhang, and Krstic}]{qi2024nofeedback}
Qi, J., Hu, J., Zhang, J., and Krstic, M. (2024{\natexlab{a}}).
\newblock Neural operator feedback for a first-order {PIDE} with spatially-varying state delay.
\newblock \emph{arXiv preprint arxiv:2412.08219}.

\bibitem[{Qi et~al.(2024{\natexlab{b}})Qi, Zhang, and Krstic}]{qi2024neural}
Qi, J., Zhang, J., and Krstic, M. (2024{\natexlab{b}}).
\newblock Neural operators for pde backstepping control of first-order hyperbolic {PIDE} with recycle and delay.
\newblock \emph{Systems \& Control Letters}, 185, 105714.

\bibitem[{Quartz et~al.(2024)Quartz, Zhou, De~Sterck, and Liu}]{quartz2024stochastic}
Quartz, T., Zhou, R., De~Sterck, H., and Liu, J. (2024).
\newblock Stochastic reinforcement learning with stability guarantees for control of unknown nonlinear systems.
\newblock \emph{arXiv preprint arXiv:2409.08382}.

\bibitem[{Schulman et~al.(2017)Schulman, Wolski, Dhariwal, Radford, and Klimov}]{schulman2017proximal}
Schulman, J., Wolski, F., Dhariwal, P., Radford, A., and Klimov, O. (2017).
\newblock Proximal policy optimization algorithms.
\newblock \emph{arXiv preprint arXiv:1707.06347}.

\bibitem[{Song et~al.(2023)Song, Romero, M{\"u}ller, Koltun, and Scaramuzza}]{song2023reaching}
Song, Y., Romero, A., M{\"u}ller, M., Koltun, V., and Scaramuzza, D. (2023).
\newblock Reaching the limit in autonomous racing: Optimal control versus reinforcement learning.
\newblock \emph{Science Robotics}, 8(82), eadg1462.

\bibitem[{Yu and Zhao(2022)}]{yu2022deep}
Yu, H. and Zhao, X. (2022).
\newblock Deep reinforcement learning with reward design for quantum control.
\newblock \emph{IEEE Transactions on Artificial Intelligence}, 5(3), 1087--1101.

\bibitem[{Zhang and Qi(2021)}]{zhang2021compensation}
Zhang, J. and Qi, J. (2021).
\newblock Compensation of spatially-varying state delay for a first-order hyperbolic {PIDE} using boundary control.
\newblock \emph{Systems \& Control Letters}, 157, 105050.

\bibitem[{Zhang and Qi(2024)}]{ZHANG2024105964}
Zhang, J. and Qi, J. (2024).
\newblock Corrigendum to “{C}ompensation of spatially-varying state delay for a first-order hyperbolic {PIDE} using boundary control”[syst. control lett. 157 (2021) 105050].
\newblock \emph{Systems \& Control Letters}, 105964.

\end{thebibliography}
                                                               
\end{document}